\pdfoutput=1
\documentclass[11pt]{article}

\usepackage[preprint]{acl}

\usepackage{bm}

\usepackage{CJKutf8}
\usepackage{makecell}
\usepackage{graphicx}
\usepackage{amsmath}
\usepackage{amssymb}
\usepackage{algorithm, algpseudocode}
\usepackage{hyperref}
\usepackage{multirow}
\usepackage{booktabs}
\algrenewcommand\algorithmicrequire{\textbf{Input:}}
\algrenewcommand\algorithmicensure{\textbf{Output:}}
\usepackage{multicol}
\usepackage{xcolor}
\usepackage{tcolorbox}
\definecolor{rq}{HTML}{1B365C}
\definecolor{rqBack}{HTML}{9ECBF7}

\newcommand{\highlightllm}[1]{\colorbox{gray!20}{\(\displaystyle #1\)}}

\usepackage{times}
\usepackage{latexsym}

\usepackage[T1]{fontenc}
\usepackage[utf8]{inputenc}

\usepackage{microtype}

\usepackage{inconsolata}

\usepackage{graphicx}

\title{Dial-In LLM: Human-Aligned LLM-in-the-loop Intent Clustering for Customer Service Dialogues}

\author{Mengze Hong$^{1}$\thanks{Work was partially done during internship at WeBank.}, Wailing Ng$^{1}$, Chen Jason Zhang$^{1}$, Yuanfeng Song$^{2}$, Di Jiang$^{1}$\thanks{Corresponding Author}\\ 
$^{1}$Hong Kong Polytechnic University, $^{2}$AI Group, WeBank Co., Ltd\\
}

\begin{document}
\maketitle

\begin{abstract}

Discovering customer intentions is crucial for automated service agents, yet existing intent clustering methods often fall short due to their reliance on embedding distance metrics and neglect of underlying semantic structures. To address these limitations, we propose an \textbf{LLM-in-the-loop (LLM-ITL)} intent clustering framework, integrating the language understanding capabilities of LLMs into conventional clustering algorithms. Specifically, this paper (1) examines the effectiveness of fine-tuned LLMs in semantic coherence evaluation and intent cluster naming, achieving over 95\% accuracy aligned with human judgments; (2) designs an LLM-ITL framework that facilitates the iterative discovery of coherent intent clusters and the optimal number of clusters; and (3) introduces context-aware techniques tailored for customer service dialogue. Since existing English benchmarks lack sufficient semantic diversity and intent coverage, we further present a comprehensive Chinese dialogue intent dataset comprising over 100k real customer service calls with 1,507 human-annotated clusters. The proposed approaches significantly outperform LLM-guided baselines, achieving notable improvements in clustering quality, cost efficiency, and downstream applications. Combined with several best practices, our findings highlight the prominence of LLM-in-the-loop techniques for scalable dialogue data mining.
\end{abstract}

\section{Introduction}

\begin{figure}[!t]
\centering
\includegraphics[width=0.9\linewidth]{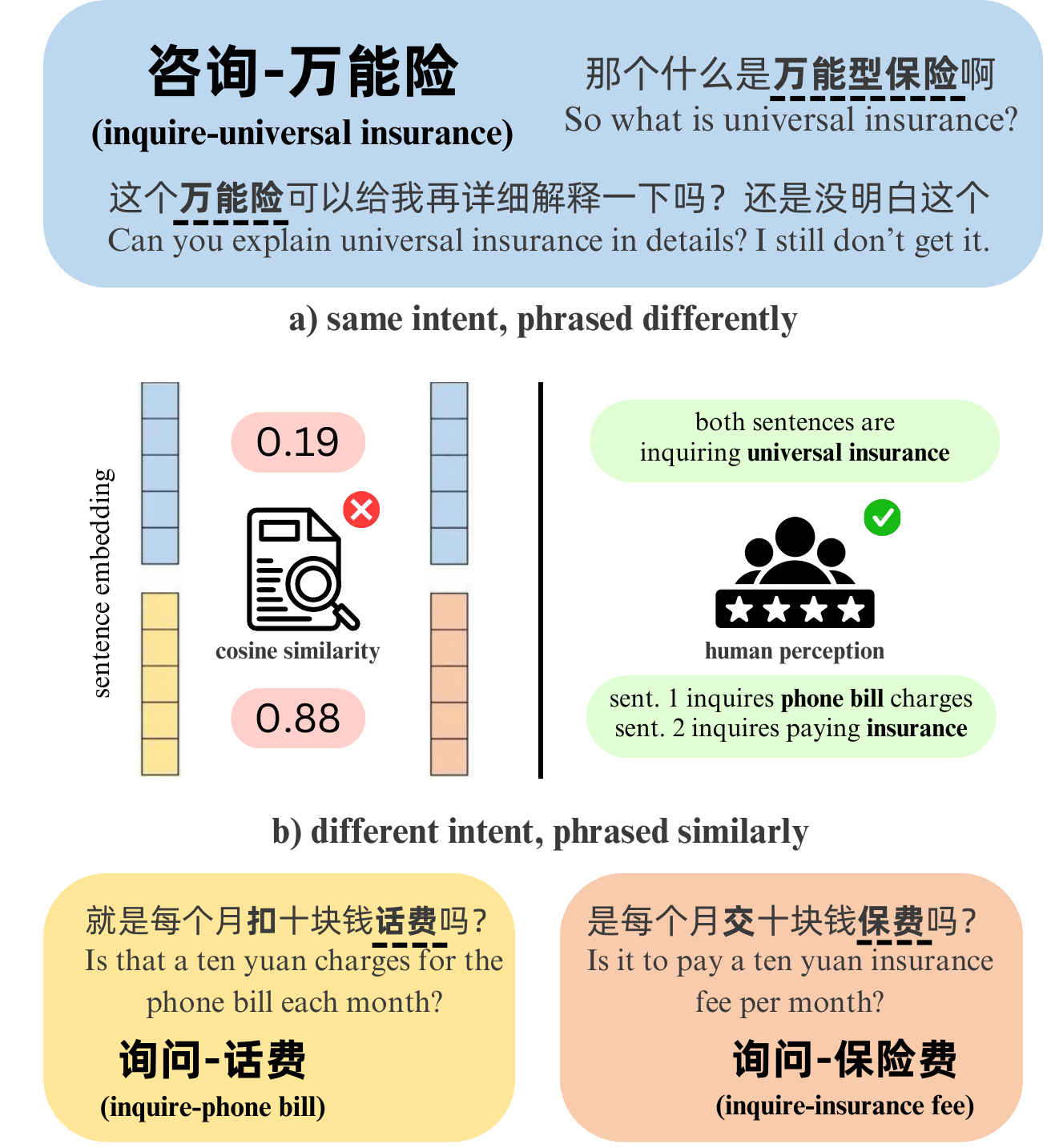}
\caption{Comparison of embedding-distance metric and human perception: cosine similarity failed to identify the same intentions under diverse expression (top), and to distinguish distinct intentions under similar expressions (bottom).}
\label{fig:similarity}
\end{figure} 

Intent discovery is a crucial task in NLP applications, such as dialogue system design \cite{hengst-etal-2024-conformal}, information retrieval \cite{jiang2016query}, and utterance pattern analysis \cite{10.1007/978-981-13-7403-6_9}. While intent clustering techniques are extensively studied to automatically identify thematic relationships within text corpora \cite{gung-etal-2023-intent}, existing research primarily focuses on developing meaningful sentence representations and relies on embedding distance metrics for optimization \cite{yin2021representationlearningshorttext, 10.1007/978-981-96-0348-0_21}. This approach often overlooks the distinctive characteristics of textual information, such as linguistic patterns and semantic diversity (see Figure \ref{fig:similarity}), and thus restricts human-aligned evaluation and validation of clustering performance \cite{vinh2009information}. This issue is particularly pronounced in languages with rich semantics, such as Chinese, where two seemingly similar sentences can convey entirely different meanings \cite{chen1993comparison, jiang2016cross}.  

With growing research interest in integrating large language models (LLMs) into the problem-solving pipeline \cite{hongposition}, LLM-guided clustering techniques have emerged, demonstrating superior performance over traditional machine learning algorithms \cite{zhang2023clusterllm}. While these methods effectively incorporate language understanding into the clustering process, they primarily focus on data preprocessing, querying LLMs for embedding refinement or data augmentation \cite{viswanathan2023large}. This approach represents a surface-level integration, potentially missing the benefits of LLMs to contribute semantic-driven guidance within the clustering process.
    
In this paper, we introduce an \textbf{LLM-in-the-loop (LLM-ITL)} intent clustering framework, designed to facilitate the iterative discovery of coherent intent clusters from semantically diverse, large-scale dialogue datasets. Our approach effectively leverages intermediate clustering results by incorporating human-aligned LLM utilities, enabling computationally efficient and human-interpretable intent clustering. The key contributions of this work are:

\begin{enumerate}
    \item We present the largest Chinese dialogue intent clustering dataset, derived from over 100,000 real-world customer service calls across the banking, telecommunication, and insurance domains. The data is annotated into 1,507 intent clusters with high semantic diversity and a substantial inclusion of noisy, out-of-domain queries, reflecting realistic and complex customer interactions.
    \item We demonstrate the effectiveness of fine-tuned small LLMs in assessing the semantic coherence of intent clusters and providing accurate intent labels across various sampling strategies, offering cost-efficient utilities for designing LLM-in-the-loop solutions.
    \item We propose an LLM-in-the-loop intent clustering framework that effectively combines the strengths of LLMs and conventional clustering algorithms. This approach outperforms state-of-the-art baselines and excels in downstream applications with 18.46\% performance gain. Furthermore, discussions on data sampling and LLM-based crowdsourcing validate best practices for real-world deployment.
\end{enumerate}

\begin{figure*}
\centering
\includegraphics[width=1\linewidth]{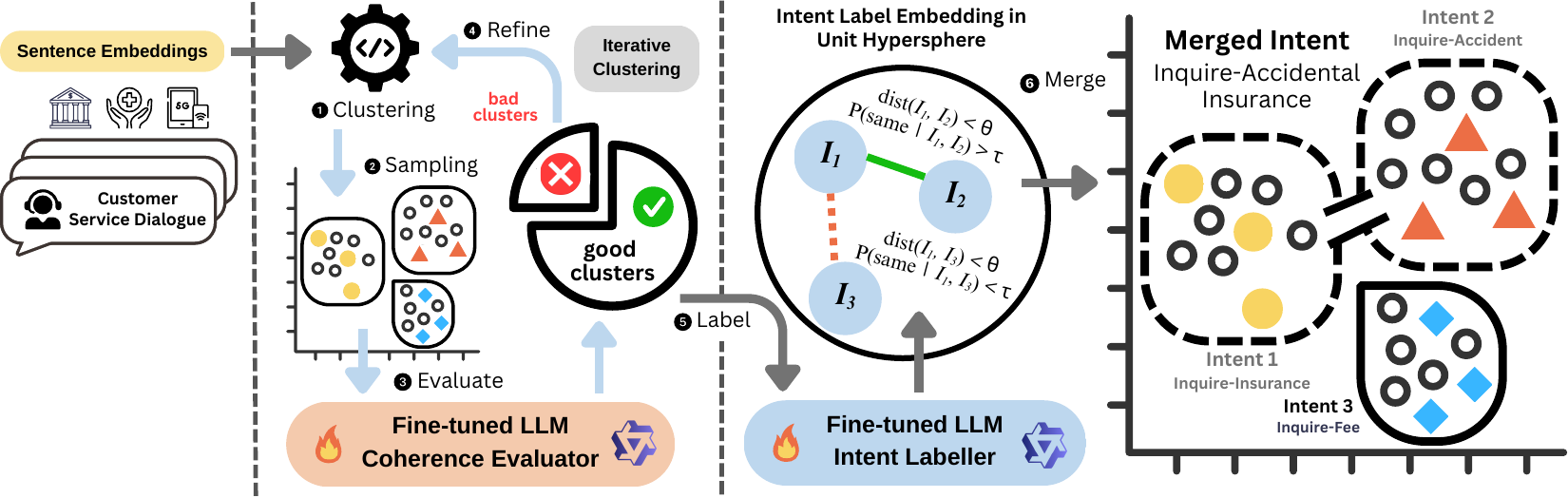}
\caption{Overview of the proposed LLM-in-the-loop framework for dialogue intent clustering.}
\label{fig:overview}
\end{figure*}

\section{Related Work}

\paragraph{LLM-guided Text Clustering.}
The integration of LLMs into text clustering has become increasingly prominent since 2023. \citet{zhang2023clusterllm} introduced ClusterLLM, an innovative approach that leverages instruction-tuned LLMs like ChatGPT to refine sentence embedding spaces through pairwise preference questions, aligning clustering granularity with user preferences. \citet{viswanathan2023large} highlighted the enhancement of clustering quality by LLMs through feature improvement, the imposition of constraints during clustering, and post-correction processes. Additionally, \citet{feng-etal-2024-llmedgerefine} proposed refining edge points with LLMs, which led to notable performance gains. 

Recently, \citet{hongposition} introduced the concept of LLM-in-the-loop machine learning, categorizing integration strategies into data-, model-, and task-level approaches. This paradigm parallels the human-in-the-loop framework \cite{chen-etal-2024-humans}, enabling LLMs to replicate human expertise and thereby enhance conventional problem-solving workflows in a cost-efficient and flexible manner. Within this taxonomy, existing work on LLM-guided clustering has primarily focused on data-centric aspects, with relatively limited exploration of modeling and task-solving aspects. This gap constrains a broader understanding of the potential of LLMs in tackling long-standing challenges in text clustering, such as improving cluster interpretability and moderating the clustering process \cite{tan1999text, jiang2021familia}.

\paragraph{Intent Clustering.} 

Intent clustering extends conventional text clustering by incorporating contextual cues and domain knowledge to uncover meaningful user intentions \cite{allahyari2017brief, hong2024neuralbayesianprogramlearningfewshot}. It serves as a fundamental step in intent induction \cite{chandrakala2024intent} and in preparing data for training intent classifiers \cite{gung-etal-2023-intent}. While conceptually related to topic modeling \cite{jiang2015teii, jiang2023probabilistic}, which seeks to uncover latent semantic themes, intent clustering tackles the more challenging task of differentiating texts that may appear lexically similar but convey contextually distinct meanings, thereby uncovering the underlying communicative goals behind sentences \cite{carberry1988modeling, qu2018analyzing}. This finer granularity makes intent clustering more demanding as a data mining task and a critical component in applications such as search engines \cite{jiang2013panorama, jiang2016query} and dialogue systems \cite{qin-etal-2023-end, hong-etal-2025-dialogue}.

While semi-supervised approaches \cite{kumar2022intent} and deep learning methods \cite{Lin_Xu_Zhang_2020, zhang2021discovering} have substantially advanced intent clustering, recent attempts in LLM-based systems highlight their practical advantages \cite{liang2024synergizing}. For instance, the IDAS method leverages in-context learning to generate descriptive utterance labels and improve sentence embeddings \cite{de2023idas}. However, challenges such as high computational cost and limited model robustness remain unresolved \cite{song-etal-2023-large}, and existing methods are evaluated based on small-scale datasets with limited complexity \cite{casanueva-etal-2020-efficient}, leading to an inadequate understanding of their effectiveness in real-world applications.

\paragraph{Highlights.} Previous studies have primarily focused on enhancing input text representations and the direct use of LLMs for cluster modification, imposing high computational cost and uncertainty. In contrast, this paper implements human-aligned LLM utilities to create an LLM-in-the-loop framework with emphasis on the intermediate clustering results. Additionally, this work releases a complex dataset from real customer service call transcriptions, enabling more practical insights and meaningful evaluations for future intent clustering.

\section{Proposed Methods}
In this section, we first outline the design of human-aligned LLM utilities. Then, we introduce a comprehensive LLM-in-the-loop intent clustering framework (see Figure \ref{fig:overview}) and show how each LLM utility contributes to the clustering process.

\subsection{LLM Utilities}

\paragraph{Coherence Evaluation.} Inspired by human behaviors in perceiving texts and making comparisons based on semantic meaning rather than surface-level work similarity \cite{doi:10.1080/01638539809545029}, this paper proposes \textbf{semantic coherence} as a more effective metric and optimization objective, measuring the semantic consistency within a cluster. The coherence evaluation is formulated as a binary classification problem, enabling intuitive interpretation and simplifying the training of the LLM evaluator (see Section \ref{sec: eval convention} for alternative formulations). \textbf{Good} clusters consist of sentences focused on a specific topic, whereas \textbf{Bad} clusters contain inconsistent or ambiguous intentions (see Table \ref{tab:Examples of GOOD/BAD Text Clusters} for examples). This task is particularly challenging for traditional machine learning classifiers due to their lack of semantic understanding \cite{10.5555/2145432.2145462}, thus necessitating fine-tuned LLMs for robust evaluation \cite{gu2025surveyllmasajudge}.

\paragraph{Cluster Naming.} Giving each cluster a concise and meaningful name is essential for many downstream applications \cite{pattnaik2024improving, luo2024llm}. In this paper, we introduce a novel naming convention, "Action-Objective," which is particularly effective for capturing dialogue intents that are typically topic-oriented (e.g., insurance, loan) and involve distinct actions (e.g., inquire, confirm). Examples of human annotations are shown in Table \ref{tab:Cluster Naming}, and a comparison of different naming conventions is presented in Table \ref{tab:naming convention} to demonstrate the effectiveness of the proposed approach.

\subsection{LLM-in-the-loop Iterative Intent Clustering with Coherence Evaluation}

At iteration \( t \), the current set of unassigned sentences is denoted as \( \mathcal{S}^{(t)} \). A special case in the first iteration, where \( \mathcal{S}^{(0)} =  \mathcal{S} \) represents the entire set of unique sentences derived from the dialogue corpus. For each candidate cluster number \( n_i \in N \), we compute:
\[
\mathcal{C}_{n_i}^{(t)} = F(\mathcal{E}^{(t)}, n_i),
\]
where \( F \) is a clustering function (e.g., K-means clustering), and \( \mathcal{E}^{(t)} = \{ \mathbf{e}_s \mid s \in \mathcal{S}^{(t)} \} \) represents the sentence embeddings. This results in $|N|$ distinct cluster assignments as the initial outcome.

Then, the semantic coherence of each cluster is evaluated using a fine-tuned LLM $\mathcal{M}_{\text{eval}}$:
\[
\mathbf{g}_{n_i}^{(t)} = \left[ \mathcal{M}_{\text{eval}}(C_1^{(t)}), \ldots, \mathcal{M}_{\text{eval}}(C_{n_i}^{(t)}) \right],
\]
where \( \mathcal{M}_{\text{eval}}(C_k) = 1 \) if the cluster is coherent (i.e., ``good'' cluster), else 0.

While the number of clusters is often known in benchmark evaluations, determining the optimal number in a noisy text corpus is both challenging and essential. Here, we propose a local search heuristic that maximizes the ``good/bad'' ratio at each iteration. The optimal \( n_{t}^* \) is given by:
\[
n_{t}^* = \arg\max_{n_i \in N} \frac{ \sum_{j=1}^{n_i} \mathbb{I} \left( \mathbf{g}_{n_i}^{(t)}[j] = 1 \right) }{ \sum_{j=1}^{n_i} \mathbb{I} \left( \mathbf{g}_{n_i}^{(t)}[j] = 0 \right) + 1},
\]
representing the best cluster number at iteration $t$. This approach enables the automatic discovery of suitable cluster numbers in a step-by-step manner. The search space $\bm{N}$ should be selected carefully to balance accuracy and efficiency, and a search space pruning strategy is proposed in Section~\ref{subsec:pruning} to enhance the searching process.

Finally, the ``good'' clusters in the optimal $\mathcal{C}_{n^*}^{(t)}$ are retained, and the remaining sentences will be refined in the next (i.e., $t+1$) iteration, enabling the iterative discovery of high-quality clusters. The proposed method is summarized in \textbf{Algorithm \ref{alg:iterative algorithm}}, and the LLM integration is highlighted.

\begin{algorithm}[t]
\caption{LLM-in-the-loop Intent Clustering}
\label{alg:iterative algorithm}
\small
\begin{algorithmic}[1]
\Require{Unlabeled sentence corpus \( \mathcal{S} \), embedding function \( f_{\text{emb}} \), coherence evaluator \( \mathcal{M}_{\text{eval}} \), candidate cluster numbers \( N = \{ n_1, \dots, n_k \} \), threshold \( \epsilon > 0 \), max iterations \( T_{\text{max}} \)}
\Ensure{Set of clusters \( \mathcal{C} \)}
\vspace{0.5em}
\State \( \mathcal{E} \gets \{ f_{\text{emb}}(s) \mid s \in \mathcal{S} \} \)
\State \( \mathcal{S}^{(0)} \gets \mathcal{S} \), \( \mathcal{C} \gets \emptyset \), \( t \gets 0 \)
\While{\( \frac{|\mathcal{S}^{(t)}|}{|\mathcal{S}|} > \epsilon \) and \( t < T_{\text{max}} \)}
    \State \( \mathcal{E}^{(t)} \gets \{ f_{\text{emb}}(s) \mid s \in \mathcal{S}^{(t)} \} \)
    \For{each \( n_i \in N \)}
        \State \( \mathcal{C}_{n_i}^{(t)} \gets F(\mathcal{E}^{(t)}, n_i) \)
        \vspace{0.2em}
        \State \highlightllm{\mathbf{g}_{n_i}^{(t)} \gets [\mathcal{M}_{\text{eval}}(C_{1}^{(t)}), \dots, \mathcal{M}_{\text{eval}}(C_{n_i}^{(t)})]}
        \vspace{0.2em}
    \EndFor
    \State \( n^* \gets \arg\max_{n_i \in N} \frac{ \sum_{j=1}^{n_i} \mathbb{I} \left( \mathbf{g}_{n_i}^{(t)}[j] = 1 \right) }{ \sum_{j=1}^{n_i} \mathbb{I} \left( \mathbf{g}_{n_i}^{(t)}[j] = 0 \right) + 1 } \)
    \State \( \mathcal{C}_{\text{good}}^{(t)} \gets \{ C_j \in \mathcal{C}_{n^*}^{(t)} \mid \mathbf{g}_{n^*}^{(t)}[j] = 1 \} \)
    \State \( \mathcal{C} \gets \mathcal{C} \cup \mathcal{C}_{\text{good}}^{(t)} \)
    \State \( \mathcal{S}^{(t+1)} \gets \mathcal{S}^{(t)} \setminus \bigcup_{C \in \mathcal{C}_{\text{good}}^{(t)}} C \)
    \State \( t \gets t + 1 \)
\EndWhile
\State \Return \( (\mathcal{C}) \)
\end{algorithmic}
\end{algorithm}

\subsection{Post-Correction with LLM-Generated Intent Labels}

Preliminary results in Table \ref{tab:similar intent} suggest that, in later iterations, the diminishing size of the unassigned sentence set \( \mathcal{S}^{(t)} \) may lead to the formation of multiple clusters capturing similar intents, resulting in smaller and less representative clusters. The embedding distances between sentences within clusters limit the natural consolidation of clusters with similar intents but different expressions \cite{khan2020sentence}, necessitating a post-correction step to merge semantically aligned clusters. Previous methods typically address this by issuing direct LLM queries to validate individual cluster assignments \cite{viswanathan2023large, feng-etal-2024-llmedgerefine}, an approach that is both computationally expensive and prone to inconsistency. In contrast, we propose a \textbf{context-aware approach}, leveraging LLMs' naming utility to robustly merge clusters based on their generated intent labels. 

At the end of the iterative clustering process, each cluster \( C_k \) receives an intent label from the fine-tuned LLM naming utility:
\[
l_k = \mathcal{M}_{\text{name}}(C_k).
\]
These labels concisely summarize the semantic information of each cluster and are mapped to the semantic space of sentence embeddings using the embedding function \( f_{\text{emb}} \):
\[
\mathbf{l}_k = f_{\text{emb}}(l_k) \in \mathbb{R}^d, \quad \| \mathbf{l}_k \|_2 = 1.
\]
This normalization positions label embeddings \( \mathbf{l}_k \) on the unit hypersphere \( \mathbb{S}^{d-1} = \{ \mathbf{x} \in \mathbb{R}^d \mid \| \mathbf{x} \|_2 = 1\} \), enabling accurate measurement of semantic relationships along the sphere’s surface rather than through straight-line distances \cite{fletcher2004principal}, thus facilitating precise intent similarity comparisons in high-dimensional space.

Then, the clusters are structured into a semantic affinity graph \( G = (V, E) \) to model relationships based on label similarity. Vertices \( V = \{ C_1, C_2, \ldots, C_K \} \) represent clusters, and the edges \( E \subseteq V \times V \) are determined using hyperspherical geometry by computing the geodesic distance, which captures angular separation between label embeddings \cite{fletcher2004principal}:
\[
\text{Dist}(\mathbf{l}_i, \mathbf{l}_j) = \arccos(\langle \mathbf{l}_i, \mathbf{l}_j \rangle),
\]
forming edges if \( \text{Dist}(\mathbf{l}_i, \mathbf{l}_j) < \theta \), with \( \theta  \) is a predefined threshold (\( \theta = 0.8 \)).

For each derived edge, a probabilistic criterion is designed to enhance the robustness of merging decisions, naturally modeling label embeddings as samples from a mixture of von Mises-Fisher distributions \cite{banerjee2005clustering}:
\[
p(\mathbf{l}_k \mid \boldsymbol{\mu}_m, \kappa_m) = Z_d(\kappa_m) \exp(\kappa_m \langle \mathbf{l}_k, \boldsymbol{\mu}_m \rangle),
\]
where \( \boldsymbol{\mu}_m = \mathbf{l}_m \in \mathbb{S}^{d-1} \) is the mean direction of the \( m \)-th intent embedding, \( \kappa_m > 0 \) controls the distribution’s tightness, and the normalization constant \( Z_d(\kappa_m) \) ensures the density integrates to 1 over the hypersphere:
\small \begin{align*}
    Z_d(\kappa) = \frac{\kappa^{d/2 - 1}}{(2\pi)^{d/2} I_{d/2 - 1}(\kappa)}.
\end{align*}
\normalsize
Thus, the probability that clusters \( C_i \) and \( C_j \) share the same intent is computed for each edge as:
\small\[
P(\text{same} \mid \mathbf{l}_i, \mathbf{l}_j) = \sum_{m=1}^K \pi_m p(\mathbf{l}_i \mid \boldsymbol{\mu}_m, \kappa_m) p(\mathbf{l}_j \mid \boldsymbol{\mu}_m, \kappa_m),
\]\normalsize
with \( \pi_m = 1/K \) as uniform mixture weights. An edge is retained only if this probability exceeds a predefined threshold \( \tau \) (e.g., \( \tau = 0.7 \)), minimizing inappropriate merges by accounting for uncertainty.

Finally, clusters linked in the affinity graph are consolidated into connected components, forming a refined cluster assignment \( \mathcal{C}' = \{ C'_1, C'_2, \ldots, C'_{K'} \} \) that eliminates redundancy, enhances interpretability, and maintains robust semantic alignment. Each merged cluster \( C'_k \) receives a new intent label reflecting its semantic content.

\subsection{Context-Aware Role Separation with LLM-Generated Intent Labels}

From a practical perspective, dialogues are often accompanied by domain-specific features or use-case scenarios (e.g., customer service calls, group discussions). This motivates the incorporation of contextual information into the clustering process \cite{ding2025retrieveandverifytablecontextselection}. In particular, the customer service dialogues typically involve only the customer and the service agent \cite{lin-etal-2022-roles}. The classification task that assigns a sentence to its associated role is relatively simple with labeled training data. To provide an unsupervised solution, we propose that the sentence roles can be naturally determined based on the LLM-generated intent label. 

By heuristics, sentences within clusters labeled as ``inquire-'' or similar actions are mostly sent from the customer, and vice versa. A two-step approach is proposed: initially, intermediate clustering results are obtained using previous approach, denoted as \( C_{\text{inter}} = \{ C_1, C_2, \ldots, C_K \} \), and these results are divided based on intent labels \( l_k \), forming two groups with distinct roles, \( R_{\text{customer}} \) and \( R_{\text{agent}} \), where:
\[
\begin{array}{c}
R_{\text{customer}} = \{ s \in S \mid l_k(s) \in \{ \text{"inquire-"}, \dots\} \}, \\
R_{\text{agent}} = \{ s \in S \mid l_k(s) \in \{ \text{"answer-"}, \dots \} \}.
\end{array}
\]
In the second step, the sentences corresponding to each role are clustered again, denoted as \( C'_{\text{customer}} = \text{Cluster}(R_{\text{customer}}) \) and \( C'_{\text{agent}} = \text{Cluster}(R_{\text{agent}}) \). Finally, the resulting customer and agent clusters are merged to produce a refined clustering that maintains a clear separation of intents by role, serving either as the final output or as an improved intermediate stage for further processing.

\section{Experiment}

\subsection{Dataset}

The proposed dataset contains 1,507 high-quality intent clusters manually annotated from over 100,000 realistic customer service calls, comprising 55,085 distinct sentences with an average length of 17 Chinese characters per sentence (see Appendix \ref{appendix:annotation} for annotation details)\footnote{Data is available at \href{https://github.com/mengze-hong/Dial-in-LLM}{GitHub} repository.}. Among these intents, 885 are identified as domain-specific (e.g., inquire-insurance), primarily concentrated within the banking, telecommunications, and insurance industries, with a focus on the Chinese context \cite{hong2025qualbench}. The remaining 622 clusters are categorized as out-of-domain (e.g., provide-location), representing general queries that commonly arise in customer service interactions. 

Compared with existing intent clustering benchmarks such as BANK77 \cite{casanueva-etal-2020-efficient} and CLINC150 \cite{larson-etal-2019-evaluation}, the proposed dataset is the first Chinese benchmark for customer service intent clustering and the largest of its kind in both the number of sentences and the number of clusters (see Table~\ref{tab:dataset description}). It presents several new challenges, including the lexical sparsity in short-text sentences, the dependence on contextual knowledge, and the difficulties in balancing accuracy with computational efficiency due to the excessively large intent size. The semantic diversity, calculated based on the average cosine distance between individual sentences and the cluster centroid \cite{casanueva-etal-2022-nlu}, depicts the complexity of this dataset and motivates the development of semantic-guided approaches.

\begin{table}[!t]
\centering \footnotesize
% \resizebox{\textwidth}{!}{
\begin{tabular}{l ccc}
\toprule
{} & \textbf{Number of}  & \textbf{Number of}  & \textbf{Semantic} \\
\textbf{Dataset} & \textbf{sentences}  & \textbf{intents} &  \textbf{diversity} \\ \midrule
\textsc{BANK77} & {3080} & {77}  & {0.209} \\
\textsc{NLU++} & {3,080} & {62}  & {0.367} \\
\textsc{CLINC(I)} & {4,500} & {150}  & {0.275} \\
\textsc{MTOP(I)} & {4,386} & {102}  & {0.234} \\
\textsc{Massive(I)} & {2,974} & {59}  & {0.351} \\ \midrule
\textbf{ours} & {\textbf{55,085}} & {\textbf{1507}}  & {\textbf{0.538}} \\
\bottomrule
\end{tabular}
% }
\caption{Comparison of the proposed customer service intent clustering dataset with existing benchmarks.}
\label{tab:dataset description}
% \vspace{-1.2em}
\end{table}

\subsection{Metrics}

\vspace{1ex}

\noindent \textbf{Normalized Mutual Information (NMI)} measures the degree of similarity between ground-truth and predicted clusters, ranging from 0 (no mutual information) to 1 (perfect correlation). However, recent work has demonstrated that NMI exhibits biased behavior, particularly in favor of larger numbers of clusters \cite{Mahmoudi2024ProofOB}, highlighting the need for complementary evaluation metrics that better reflect human-perceived clustering quality.

\vspace{1ex}

\noindent \textbf{Goodness score} measures the proportion of good clusters evaluated by the fine-tuned LLM evaluator. For intermediate steps of iteration where the number of clusters is unknown, the good/bad ratio is used as an invariant measure of clustering quality:
\[
\text{goodness}_i = \frac{\text{\# good clusters}}{\text{\# bad clusters}}
\]
\noindent For the final clustering results, the percentage of good clusters among all clusters is reported. Note that the evaluation uses a different LLM evaluator than the one used for intermediate evaluation to ensure the fairness of the reported metric:
\[
\text{goodness}_{final} = \frac{\text{\# good clusters}}{\text{\# \textbf{total clusters}}}
\]

\noindent This metric offers a comprehensive understanding of the quality of the produced intent clusters, allowing for accurate cluster-level evaluation. Additionally, it can be easily deployed in applications without ground-truth annotations, thereby eliminating the need for human involvement.

\subsection{Implementations}

Four open-sourced Chinese LLMs\footnote{The models are   available at: \href{https://huggingface.co/Qwen/Qwen2.5-7B}{Qwen2.5-7B}; \href{https://huggingface.co/Qwen/Qwen2.5-14B}{Qwen2.5-14B}; \href{https://huggingface.co/baichuan-inc/Baichuan2-7B-Base}{Baichuan2-7B  }; \href{https://huggingface.co/THUDM/chatglm3-6b}{ChatGLM3-6b}} are fine-tuned using LoRA \cite{Hu2021LoRALA} on 4 $\times$ Nvidia A100 GPUs. For coherence evaluation, a human-annotated training dataset of 1,772 intent clusters labeled as ``good'' or ``bad'' is used. For cluster naming, a training dataset of 2,500 clusters, each containing 20 sentences, is annotated with the ``Action-Objective'' labels. Semantic-rich embeddings are generated using the BAAI General Embeddings (BGE) model\footnote{\href{https://huggingface.co/BAAI/bge-large-zh-v1.5}{https://huggingface.co/BAAI/bge-large-zh-v1.5}} \cite{bge_embedding}. The best-performing clustering algorithm, hierarchical clustering, and LLM utilities, \textit{qwen14b} and \textit{chatglm3-6b}, are selected for LLM-in-the-loop clustering. Additionally, we employ \textit{random} and \textit{convex} sampling to select 20 representative sentences per cluster as LLM inputs, reducing computational overhead while preserving intent coverage.

\section{Results and Discussions}

\begin{CJK}{UTF8}{gkai}
\begin{table}[!t]
    \centering
    \renewcommand\arraystretch{2} 
    \resizebox{\columnwidth}{!}{
    \begin{tabular}{|c|c|c|c|c|c|c|c|c|c|c|c|c|c|} \hline
        \textbf{LLM} &\thead{\textbf{qwen7b}}&\thead{\textbf{qwen14b}}&\thead{\textbf{baichuan2-7b}}&\thead{\textbf{chatglm3-6b}}\\ \hline
       Accuracy &96.25\%&\textbf{97.50\%}&89.17\%&95.83\%\\ \hline
    \end{tabular}}
    \caption{Performance of fine-tuned LLMs in evaluating semantic coherence of intent clusters.}
    \label{tab:llm goodness}
    \vspace{-0.7em}
\end{table}
\end{CJK}

\begin{CJK}{UTF8}{gkai}
\begin{table}[!t]
    \centering
    \renewcommand\arraystretch{2} 
    \resizebox{\columnwidth}{!}{
    \begin{tabular}{|c|c|c|c|c|c|c|c|c|c|c|c|c|c|} \hline
        \textbf{LLM} &\thead{\textbf{qwen7b}}&\thead{\textbf{qwen14b}}&\thead{\textbf{baichuan2-7b}}&\thead{\textbf{chatglm3-6b}}\\ \hline
       Accuracy & 92.8\% & 94.3\%&94.3\%  &\textbf{94.4\%} \\ \hline
    \end{tabular}}
    \caption{Performance of fine-tuned LLMs in naming intent clusters.}
    \label{tab:llm naming}
    \vspace{-1em}
\end{table}
\end{CJK}

\subsection{LLM Utilities}
\label{sec: utilities}

Table \ref{tab:llm goodness} presents the performance of the fine-tuned LLM coherence evaluator tested on 480 unseen clusters. The results indicate that mainstream open-source LLMs can effectively serve as robust evaluators for assessing the quality of intent clusters and providing human-aligned judgments. For cluster naming, since labels are not unique, accuracy is manually evaluated by four human experts based on alignment with the true labels in the dataset. The results in Table \ref{tab:llm naming} demonstrate that the fine-tuned LLMs show promising performance in generating intuitive names and adhering strictly to the predefined ``Action-Objective'' format.

\begin{table*}[!t]
\centering \footnotesize
\resizebox{0.95\textwidth}{!}{
\begin{tabular}{l llll}
\toprule
\textbf{Method} & \textbf{NMI (Mean $\pm$ Std)} & \textbf{NMI Gain} & \textbf{\#Good (Mean $\pm$ Std)} & \textbf{\#Good Gain} \\ 
\midrule

\multicolumn{5}{l}{\textsc{\textbf{\textit{Baselines}}}} \vspace{0.05cm}\\
K-Means & 0.7899 $\pm$ 0.0135 & - & 94.8\% $\pm$ 1.3\% & - \\ 
GMMs & 0.7903 $\pm$ 0.0140 & +0.05\% & 91.1\% $\pm$ 1.6\% & -3.90\% \\ 
Hierarchical & 0.8001 $\pm$ 0.0128 & +1.29\% & 94.9\% $\pm$ 1.2\% & +0.11\% \\  

\midrule
\multicolumn{5}{l}{\textsc{\textbf{\textit{LLM-Guided Clustering Baselines}}}} \vspace{0.05cm}\\
ClusterLLM \cite{zhang2023clusterllm} & 0.7284 $\pm$ 0.0168 & -7.79\% & 91.2\% $\pm$ 1.8\% & -3.80\% \\
IDAS \cite{de2023idas} & 0.8109 $\pm$ 0.0105 & +2.66\% * & 93.6\% $\pm$ 1.2\% & -1.27\% \\
Keyphrase \cite{viswanathan2023large} & 0.8371 $\pm$ 0.0098 & +5.97\% *** & 94.5\% $\pm$ 1.0\% & -0.32\% \\
LLMEdgeRefine \cite{feng-etal-2024-llmedgerefine} & 0.7411 $\pm$ 0.0155 & -6.19\% & 87.2\% $\pm$ 2.0\% & -8.02\% \\

\midrule
\multicolumn{5}{l}{\textsc{\textbf{\textit{Proposed Method (Context-Free)}}}} \vspace{0.05cm}\\
LLM-ITL (random) & 0.8202 $\pm$ 0.0095 & +3.84\% ** & 97.7\% $\pm$ 0.6\% & +3.06\% *** \\
LLM-ITL (convex) & 0.8208 $\pm$ 0.0090 & +3.92\% ** & \textbf{97.8\% $\pm$ 0.5\%} & +3.16\% *** \\
LLM-ITL + keyphrase & 0.8378 $\pm$ 0.0085 & +6.06\% *** & 96.4\% $\pm$ 0.6\% & +1.69\% * \\

\midrule
\multicolumn{5}{l}{\textsc{\textbf{\textit{Proposed Method (Context-Aware)}}}} \vspace{0.05cm}\\
LLM-ITL + merge & 0.8420 $\pm$ 0.0178 & +6.59\% ** & 97.8\% $\pm$ 1.3\% & +3.16\% *** \\
LLM-ITL + role & 0.8679 $\pm$ 0.0068 & +9.86\% *** & 97.2\% $\pm$ 0.5\% & +2.53\% ** \\
LLM-ITL + role + merge & \textbf{0.8826 $\pm$ 0.0060} & \textbf{+11.76\% ***} & 97.6\% $\pm$ 0.4\% & +2.95\% *** \\

\bottomrule
\end{tabular}}
\caption{Performance of baselines and proposed methods in dialogue intent clustering. Gains are computed relative to the K-Means baseline. The best results in each column are \textbf{bolded}.}
\label{tab:main_result}
\end{table*}

\subsection{Main Results}

Table~\ref{tab:main_result} reports the main results on the proposed dataset, with all evaluation metrics averaged over five random seeds. Among the LLM-guided clustering baselines, the data-centric keyphrase expansion method, which refines sentence embeddings via LLM-summarized keyphrases, achieved the most significant performance gain \cite{viswanathan2023large}. However, deriving pairwise constraints for fine-tuning the embedding model, as in ClusterLLM, was ineffective due to the excessive number of clusters in the dataset. Interestingly, goodness evaluation does not always align with the NMI score. For example, while keyphrase expansion improved NMI, the generated keyphrases could distort the original sentence's meaning, leading to less representative clusters and a slightly lower goodness score compared to the base model. This highlights the need to balance ground truth alignment with semantic coherence during evaluation.

The proposed LLM-in-the-loop intent clustering demonstrated satisfactory improvements in NMI and significant enhancements in the goodness score. In a context-free setting, iterative clustering with convex sentence sampling outperformed both ClusterLLM and IDAS baselines. Effective keyphrase data augmentation further improved NMI, highlighting the potential of integrating data-centric methods into the pipeline. Furthermore, the context-aware approaches consistently improved performance by incorporating dialogue roles and cluster merging, resulting in a 4.48\% increase in NMI without compromising the goodness score. This emphasizes the importance of task-centric design in leveraging intent labels effectively for the problem of intent clustering.

Notably, the proposed method does not require prior knowledge of the number of clusters and automatically performs parameter searches during clustering, unlike baselines that depend on the true cluster count. This adaptability highlights its strong application potential for extracting intent clusters from noisy text corpora~\cite{akama-etal-2020-filtering}.

\begin{table}[!t]
\centering \footnotesize
\resizebox{\columnwidth}{!}{
\begin{tabular}{lcccc}
\toprule
\textbf{Method} & \textbf{Bank77} & \textbf{CLINC(I)} & \textbf{MTOP(I)} & \textbf{Massive(I)} \\
\midrule
SCCL & 81.77 & 92.94 & 73.52 & 73.90 \\
Self-supervise & 80.75 & 93.88 & 72.50 & 72.88 \\
ClusterLLM & \textbf{85.07} & 94.00 & \textbf{73.83} & 77.64 \\
IDAS & 82.84 & 92.35 & 72.31 & 75.74 \\
LLMEdgeRefine & - & \textbf{94.86} & 72.92 & 76.66 \\ \midrule
\textbf{ours} & 82.32 & 94.12 & 72.45 & \textbf{78.12} \\
\bottomrule
\end{tabular}}
\caption{NMI (\%) performance of different clustering methods on four English intent benchmarks.}
\label{tab: english result}
\end{table}

\subsection{Evaluation on English Benchmarks}
\label{sec:eng}
To validate the effectiveness of the proposed LLM-in-the-loop technique, we evaluate the context-aware (best-performing) method on four widely used English benchmarks \cite{fitzgerald-etal-2023-massive}. The Llama-7b model is fine-tuned with 800 intent clusters (200 samples from each dataset) annotated by human experts to derive Action-Objective intent labels and bad clusters through perturbation. As shown in Table \ref{tab: english result}, our method delivers performance comparable to state-of-the-art LLM-guided methods and outperforms on Massive(I) dataset. Furthermore, the computational cost is lower than the compared methods as measured in Section \ref{sec:cost}.

\begin{table}[!t]
\centering
\resizebox{\columnwidth}{!}{
\begin{tabular}{lcl}
\toprule
\textbf{Method} & \textbf{Accuracy} & \textbf{Performance Gain} \\
\midrule
K-Means (Baseline)        & 0.65 & - \\
ClusterLLM                & 0.62 & -4.62\% \\
LLMEdgeRefine             & 0.63 & -3.08\% \\
IDAS                      & 0.68 & +4.62\% * \\
Keyphrase Expansion               & 0.72 & +10.77\% *** \\
LLM-ITL (context-free)           & 0.73 & +12.31\% *** \\
LLM-ITL (context-aware)    & \textbf{0.77} & \textbf{+18.46\%} *** \\
\bottomrule
\end{tabular}}
\caption{Accuracy of BERT classifiers trained on datasets generated by different clustering methods.}
\vspace{-0.5em}
\label{tab:bert_performance}
\end{table}

\subsection{Application Performance}
One objective of intent clustering is to build a high-quality labeled dataset for training intent classifiers that can handle future inputs \cite{gung-etal-2023-intent}. To evaluate the practical effectiveness of our methods, we trained BERT classifiers (bert-base-chinese) on clustered data generated by different approaches.

As shown in Table~\ref{tab:bert_performance}, the baseline method only achieved 65\% accuracy. ClusterLLM and LLMEdgeRefine performed slightly worse, likely due to their heavier reliance on LLM-driven modifications, which can introduce label noise or overly fine-grained distinctions that reduce cluster consistency. In contrast, IDAS applies a more conservative refinement strategy, leading to a modest but stable improvement. Our proposed LLM-in-the-loop methods substantially outperformed existing approaches: LLM-ITL reached 73\% accuracy, and its enhanced variant achieved 77\%, representing an 18.46\% relative improvement. These results are consistent with the observed cluster quality and provide additional empirical evidence that our method produces high-quality, human-aligned intent clusters with significant practical advantages.

\subsection{Analysis of Computational Cost}
\label{sec:cost}

We compare the computational cost of the proposed method with existing LLM-guided clustering on the Bank77 dataset, which contains \( S = 3,080 \) test sentences and \( N = 77 \) true clusters. Our LLM-in-the-loop approach evaluates candidate cluster numbers \( N = [10, 30, 50, 70] \), requiring \( \sum_{n_i \in N} n_i = 160 \) calls per iteration. With \( T = 3 \) iterations, this totals 480 calls, each processing 20 sentences for coherence evaluation. Cluster naming requires additional LLM calls based on the final cluster count, resulting in approximately 560 calls. In contrast, ClusterLLM uses a fixed triplet sampling strategy with \( Q = 1,024 \), resulting in a constant cost of 1,024 calls. Keyphrase expansion generates one keyphrase per sentence, totaling 3,080 calls. While our method processes more input tokens per call, it remains cost-efficient as the output is limited to a short “good” or “bad” label or an intent label.

\subsection{Ablation Studies}

\subsubsection*{Data Sampling and LLM Crowdsourcing for Effective Coherence Evaluation}
\label{sec:sample}

\begin{figure}[!t] % Use figure* to span both columns
    \centering
    % Image part
    \includegraphics[width=0.9\columnwidth]{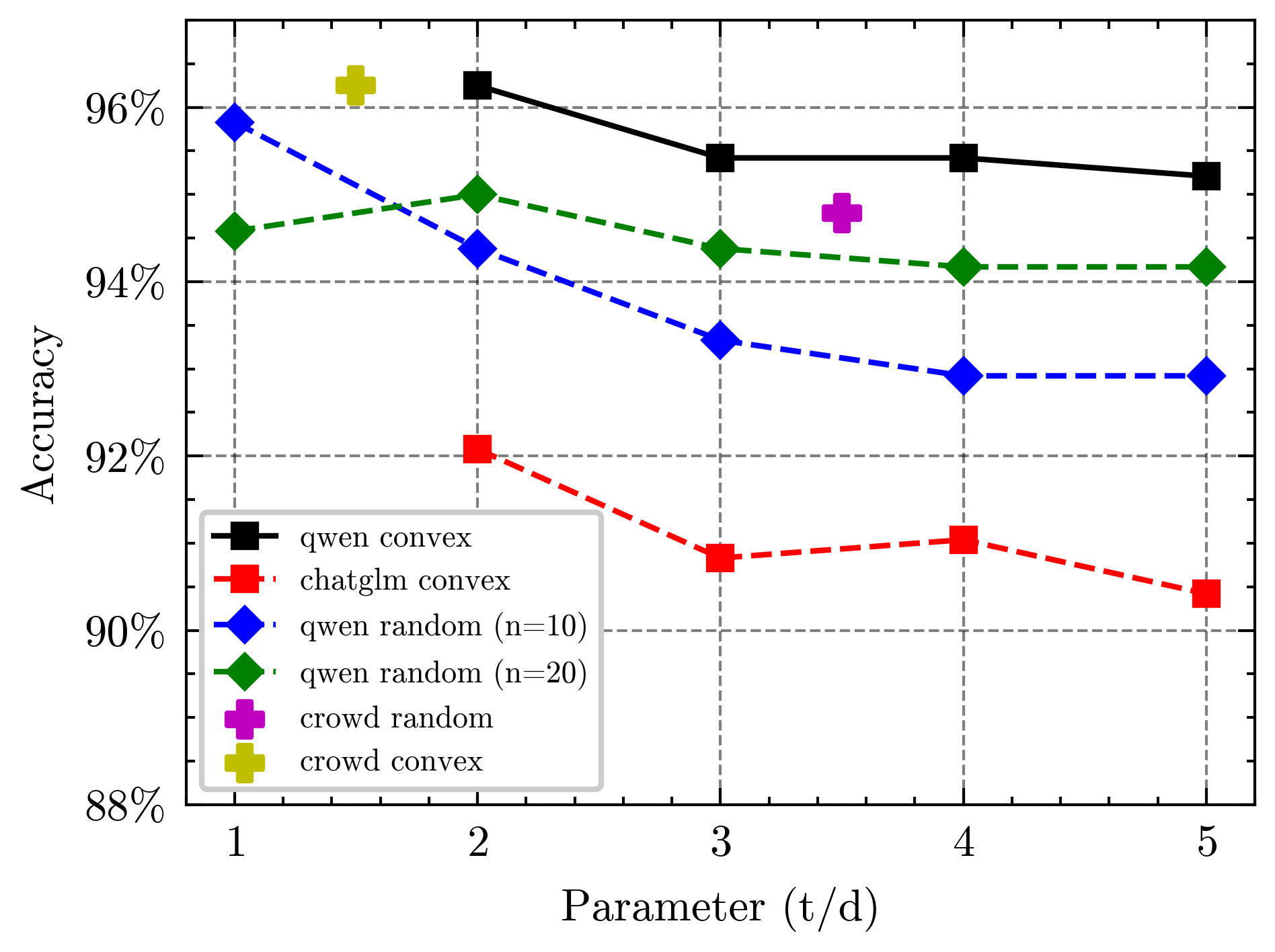} % 
    \vspace{-0.5em}
    \caption{Performance comparison of sampling methods and hyperparameters for effective LLM evaluation.}
    \label{fig:sampling}
\end{figure}

Optimizing input to the LLM evaluator is crucial for improving performance and reducing costs \cite{song-etal-2025-less, jiang-etal-2024-longllmlingua}. In this ablation study, we compare various sampling techniques to identify best practices for coherence evaluation. As shown in Figure \ref{fig:sampling}, convex sampling outperforms random sampling in selecting representative sentences. Notably, increasing either the sample repetitions \( t \) for repeated validation or the convex hull dimension \( d \) consistently reduces performance. This suggests that simpler parameter choices yield higher accuracy, a trend also seen with the ChatGLM model (red line) and summarized in Table~\ref{tab:sample_method} with numerical comparison.

To better assess the consistency and alignment between LLM judgments and human perceptions, we utilize the concept of crowdsourcing for collaborative coherence evaluation \cite{hongposition}. The proposed LLM crowd consists of diverse entities \cite{zhang2024similarity} represented by four fine-tuned LLMs and uses a majority voting mechanism to aggregate their judgments \cite{doi:10.1126/sciadv.adp1528}. Based on the evaluation, we argue that \textbf{a single fine-tuned LLM can effectively serve as a robust evaluator}, as evidenced by the comparable performance to crowdsourced accuracy (cross markers in Figure \ref{fig:sampling}) when using convex sampling.

\begin{table}[!t]
\resizebox{\columnwidth}{!}{
\begin{tabular}{l c c}
\toprule
\textbf{Method} & \textbf{NMI} & \textbf{Goodness} \\
\midrule
\textbf{Geodesic distance (probabilistic)} & \textbf{0.8420} & \textbf{97.8}\%\\
Cosine similarity (deterministic) & 0.8102 & 94.7\% \\
Cosine similarity (probabilistic) & 0.8242 & 95.2\% \\
Geodesic distance (deterministic) & 0.8309 & 95.4\% \\
\bottomrule
\end{tabular}}
\caption{Ablation study on cluster merging methods.}
\label{tab:ablation_merge}
\end{table}

\subsubsection*{Effectiveness of Cluster Merging Techniques}
\label{subsec:merge}
Table \ref{tab:ablation_merge} evaluates the effects of distance measures and merging strategies on the proposed cluster merging method. The results show that geodesic distance in hyperspherical space outperforms cosine similarity by capturing deeper semantic relationships. Furthermore, the probabilistic merging strategy consistently enhances robustness by accounting for uncertainty, leading to better clustering quality with any distance measure. These findings underscore the value of combining an appropriate distance metric with a probabilistic approach to achieve optimal merging performance.

\begin{CJK}{UTF8}{gkai}
\begin{table}[!t]
    \centering
    \renewcommand\arraystretch{2} 
    \resizebox{\columnwidth}{!}{
    \begin{tabular}{|c|c|c|c|c|c|c|c|c|c|c|c|c|c|} \hline &\thead{\textbf{Top1}}&\thead{\textbf{Top2}}&\thead{\textbf{Top3}}&\thead{\textbf{Top4}} &\thead{\textbf{Top5}}\\ \hline
       Accuracy & 26.32\% & 47.37\%& 73.68\%  & 84.21\% & 89.47\% \\ \hline
    \end{tabular}}
    \caption{Search space pruning by LLMs.}
    \label{Pruning Search Space by LLMs}
    \vspace{-1em}
\end{table}
\end{CJK}

\subsubsection*{Parameter Search Space Pruning}
\label{subsec:pruning}

In practice, the search space for the number of clusters can be extensive, ranging from a few options to several hundred. To mitigate the computational costs associated with traditional model-based selection, we propose an LLM-native method for search space pruning, which predicts the optimal cluster number for subsequent iterations using logs from previous iterations (see Table \ref{tab:log}). By pruning the search space and selecting the top 5 non-repetitive predicted solutions, we achieved an accuracy of 89.47\% (see Table~\ref{Pruning Search Space by LLMs}), indicating a high likelihood that the optimal cluster number is among predicted candidates. This approximation approach significantly reduces redundant model fittings and improves the efficiency of applying the proposed methods to large-scale datasets.

\subsubsection*{Binary Judgment vs. Numerical Scoring for Coherence Annotation}  
\label{sec: eval convention}  
Finally, to identify the optimal annotation strategy for coherence evaluation, we compared a numerical 1 -- 4 scoring system (1: very poor, 4: very coherent) with the proposed binary good/bad judgment. We annotated 1,000 clusters with input from five human experts, finalized scores through majority voting, and fine-tuned four Chinese LLMs using this dataset. The models were then evaluated on 200 additional clusters, assigning scores five times per cluster and consolidating results via majority voting (see results in Table \ref{tab:numerical judgement}). The 1 -- 4 scale demonstrated low accuracy and inconsistency, with 19\% of samples observed to have significant disagreement (e.g., three `1’s and two `4’s) in Qwen-2.5-14B, making it unreliable for assessing cluster quality or aligning with human judgment. In contrast, the simpler binary good/bad judgment provided more consistent, interpretable, and reliable quality assessments, demonstrating its superiority as an evaluation protocol.

\begin{table}[!t]
    \centering
    \renewcommand\arraystretch{2} 
    \resizebox{\columnwidth}{!}{
    \begin{tabular}{|c|c|c|c|c|c|c|c|c|c|c|c|c|c|} \hline
        \textbf{LLM} &\thead{\textbf{qwen7b}}&\thead{\textbf{qwen14b}}&\thead{\textbf{baichuan2-7b}}&\thead{\textbf{chatglm3-6b}} \\ \hline
        Accuracy & 62.50\% & \textbf{65.00\%} & 58.50\% & 60.00\% \\ \hline
    \end{tabular}}
    \caption{Performance of fine-tuned LLMs in assessing cluster coherence using a numerical 1 -- 4 scale.}
    \label{tab:numerical judgement}
    \vspace{-1em}
\end{table}

To further validate the effectiveness and robustness of our annotation protocol, we conducted an inter-annotator agreement study with five experts. They annotated 100 intent clusters generated by the K-means algorithm, covering 2,000 sentences from recent customer service dialogues. Each cluster was evaluated using (i) a binary Good/Bad judgment and (ii) a finer-grained 1 -- 4 rubric. The Fleiss’ kappa ($\kappa$) for the binary scheme reached 0.82, indicating “almost perfect” agreement~\cite{landis1977measurement}, while the 1 -- 4 rubric achieved only 0.59 (“moderate” agreement) due to variability in intermediate scores (2 and 3). These results show that binary annotation provides a more reliable and interpretable measure of cluster coherence, reinforcing the robustness of our methodology.

\section{Conclusion}
This paper tackles dialogue intent clustering through a human-aligned LLM-in-the-loop framework. Experiments on a large-scale Chinese customer service dataset demonstrate that fine-tuned LLM utilities are highly effective for semantic coherence evaluation and cluster labeling, enabling consistent improvements over existing LLM-guided baselines in both clustering quality and computational efficiency. Beyond achieving state-of-the-art performance, our study offers strong empirical evidence for the effectiveness of LLM-in-the-loop methodologies, with ablation studies highlighting best practices. Future work should refine evaluation beyond coherence to capture interpretability and expressiveness, extend the clustering framework to multilingual settings, and explore deeper task-centric integration of LLMs to further advance intent mining in real-world applications.

\section*{Limitations}

Despite the effectiveness of the proposed LLM-in-the-loop intent clustering method, this study has several limitations. First, while cluster coherence is a practical and intuitive quality indicator, its sole reliance overlooks other critical attributes, such as meaningfulness and expressiveness, which are equally important for assessing the quality of intent clusters. For example, a cluster labeled ``express-feeling'' may be too broad and could be refined into more specific clusters like ``express-appreciation'' to improve interpretability and applicability. Additionally, the binary evaluation and numerical scoring used are both deterministic metrics with fixed scales, neglecting probabilistic judgments and confidence intervals that could provide deeper assessment insights and enhance flexibility for LLM-in-the-loop integration.

Second, the proposed intent clustering dataset is limited to the Chinese language and specific domains. Although it excels in capturing large-volume, realistic customer service dialogues, its scope restricts multi-domain and multilingual generalizability, potentially limiting the applicability of findings to other languages and domains. Moreover, this paper focuses solely on intent clustering, an initial step in the broader intent discovery process. Subsequent steps, such as analyzing intent trajectories and recognizing intents at the document level rather than the sentence level, encourage further investigation with more diverse LLM utilities and varied integration strategies to better reveal the practical value of LLM-in-the-loop methodologies.

\section*{Acknowledgments}

This paper is partially supported by several ongoing projects led or coordinated by Prof. Zhang Chen, including P0045948 and P0046453 (industry donations from Accel Group Holding Limited and Minshang Creative Technology Holdings Limited), P0046701 and P0046703 (PolyU internal research funding), P0048183 and P0048191 (Research Matching Grant Scheme funded by the University Grants Committee), P0048887 (Innovation and Technology Fund - ITSP, ITS/028/22FP), P0051906 (RGC Early Career Scheme, 25600624), and P0054482 (industry donations from Two Square Capital Limited).

\bibliography{custom}

\newpage
\appendix

\section{Data Collection and Annotation}
\label{appendix:annotation}
This section details the three-stage data collection and annotation process for the proposed Chinese customer service dialogue intent clustering dataset.

\paragraph{Data Collection.} The raw dialogue dataset was derived from audio transcriptions of customer service calls in three major domains: banking, telecommunication, and insurance. In total, the dataset contained 11,879 calls. To ensure data integrity and protect confidentiality, strict filtering was applied to exclude sensitive content, resulting in 8,184 dialogues with 69,839 sentences. Further cleaning was conducted to remove duplicate sentences, yielding a final set of 55,085 unique sentences.  

\paragraph{Data Annotation.} To construct the intent clustering dataset, we recruited 15 human experts to conduct the annotations. Each annotator has over five years of professional experience in customer service call centers, with domain expertise aligned with the dataset domains: banking, telecommunication, and insurance. All annotators are fluent in Chinese, familiar with domain-specific terminology, and experienced in dialogue annotation or quality assurance tasks. Prior to annotation, a mandatory training session was held, during which the authors provided 50 intent clusters annotated with Good/Bad judgments and Action–Objective intent labels as demonstration and reference examples. This training ensured a consistent understanding of the guidelines across all annotators. The annotation process is outlined as follows:

\begin{enumerate}
    \item K-means clustering was initially applied with $n=2000$, serving as a starting point for the data annotation.
    \item Each expert assessed the initial clusters by inspecting the semantic coherence of the sentences. They were instructed: ``Label the cluster as 'Good' if all sentences share the same intent; otherwise, label it as 'Bad'. This resulted in 1283 good clusters and 717 bad clusters.
    \item For the good clusters, annotators were asked to label the underlying intent using the naming convention ``Action-Objective.''  After intent annotation, 1255 clusters with unique intentions were retained, and 28 clusters were merged due to replicated intent.
    \item For the bad clusters, annotators reassigned each sentence to the appropriate intent cluster based on the annotated labels. Sentences that did not fit into any preexisting clusters were assigned to new clusters, and the same process was repeated as in Step 2.
\end{enumerate}

\paragraph{Data Verification.} In the final stage, a separate group of 10 experts reviewed the annotated clusters for accuracy and consistency. Any discrepancies were resolved through consensus, ensuring the dataset's reliability and validity for further analysis. The final dataset consists of 1,507 high-quality intent clusters.

\section{Prompt Template}
The prompts for coherence evaluation and cluster naming are translated into English for demonstration purposes. Each input is accompanied by a few-shot demonstration with five input-output pairs to ensure consistency in the output format and enhance understanding of the task.
    
\begin{tcolorbox}[top=1pt, bottom=1pt, left=1pt, right=1pt]
  \textbf{Coherence Evaluation - }~\textit{Your are a helpful assistant for sentence clustering. Based on the relevancy and common points of the following sentences in a cluster, classify the cluster as: ``Good'' or ``Bad''. Only provide the label without any additional content.} 
  
  \vspace{1em}
  \textit{Example: input:[sentences] output:[label]}
  
  \vspace{0.5em}
  \textit{input:\{[sentences]\} output:}
\end{tcolorbox} 

\begin{tcolorbox}[top=1pt, bottom=1pt, left=1pt, right=1pt]
  \textbf{Cluster Naming - }~\textit{Your are a helpful assistant for sentence clustering. Based on the relevancy and common points of the following sentences in a cluster, summarize the cluster with an ``Action-Objective'' label. Only provide the label without any additional content.} 
  
  \vspace{1em}
  \textit{Example: input:[sentences] output:[label]}
  
  \vspace{0.5em}
  \textit{input:\{[sentences]\} output:}
\end{tcolorbox}

\begin{table}[!t]
\centering
\resizebox{\columnwidth}{!}{
\begin{tabular}{lcc}
\toprule
\textbf{Model} & \textbf{Coherence Evaluation} & \textbf{Cluster Naming} \\
\midrule
LLaMA-2-7B (LoRA)     & 94.0\% $\pm$ 1.2\%  & 89.0\% $\pm$ 1.5\% \\
LLaMA-2-13B (LoRA)    & \textbf{95.0\%} $\pm$ 1.1\%  & \textbf{93.5\%} $\pm$ 1.4\% \\
Mistral-7B (LoRA)     & 91.5\% $\pm$ 2.3\%  & 82.0\% $\pm$ 1.9\% \\
GPT-3.5 (API)         & 91.0\% $\pm$ 1.5\%  & 86.0\% $\pm$ 1.8\% \\
GPT-4 (API)           & 94.5\% $\pm$ 1.2\%  & 91.0\% $\pm$ 1.3\% \\
\bottomrule
\end{tabular}}
\caption{Performance of different LLMs on coherence evaluation and cluster naming for the English dataset.}
\label{tab:coherence_naming_accuracy}
\end{table}

\section{More Results: Evaluation with Proprietary LLM}

Given the resource-intensive nature of LLM fine-tuning, we conducted additional experiments using proprietary LLMs accessed through OpenAI APIs: GPT-3.5, GPT-4, and GPT-4o. These widely adopted models deliver strong performance across diverse tasks and do not require task-specific fine-tuning, thereby alleviating the data scarcity issue, but they also incur higher operational costs due to token consumption. On the same Chinese coherence evaluation dataset with 480 clusters, these models achieved accuracies of 89.58\%, 93.54\%, and 94.17\%, respectively. Notably, the smaller fine-tuned Qwen-2.5-7B reached 96.25\% accuracy, surpassing these advanced proprietary models while significantly reducing API-related costs. For comparison, the vanilla Qwen-2.5-7B model (without fine-tuning) obtained a much lower accuracy of 75.63\%, further underscoring the importance of fine-tuning.

On the English dataset, we report results for GPT-3.5 and GPT-4, as well as two additional fine-tuned models: LLaMA-2-13B (LoRA) and Mistral-7B (LoRA). The LLaMA and Mistral models were fine-tuned on the same 800 intent clusters as LLaMA-2-7B in Section~\ref{sec:eng} and evaluated on 100 English intent clusters annotated by human experts with Good/Bad and Action-Objective intent labels. Each model was run five times with different random seeds to ensure robust performance metrics. Results in Table~\ref{tab:coherence_naming_accuracy} show that scaling up model size (e.g., 7B vs. 13B) improves performance, while fine-tuned smaller LLMs often outperform large proprietary LLMs, consistent with the findings on the Chinese dataset. These results reinforce the value of fine-tuning and suggest that smaller, cost-efficient models can play a critical role in data mining within the LLM-in-the-loop framework.

\onecolumn

\begin{CJK}{UTF8}{gkai} 
\begin{table*}[!t] \centering
\resizebox{\textwidth}{!}{
\begin{tabular}{|c|l|l|}
\hline
\thead{\textbf{Cluster Coherence}} &  \thead{\textbf{Original Sentences}} & \thead{\textbf{English Translation}} \\ \hline
Good  & \thead{"给企业固定资产买保险，大约能投大约得投保多少呢",\\ "就是假如我有一百万的企业固定资金买保险大概投保多少钱啊",\\ "请问一下我想咨>询一下企业财产保险", \\"您好我想了解一下这个企业财产保险", \\"就了解一下这个企业财产保险","你们是财险",\\ "是哦这个属于财险了对吧", "企业财产保险的", \\"这些都属于财产险对吗", "财险人工那你这是", \\就是如果我要为我这个私营企业买这个保险需要什么手续",\\"财产险", "呃企业财产保险是以什么为保"}&\thead{"If I buy insurance for a company's fixed assets,\\ how much insurance will it cost?",\\ "That is, if I have a million corporate fixed assets, \\how much will it cost to buy insurance?",\\ "Excuse me, I would like to consult> Ask about corporate property insurance", \\"Hello, I want to know about this corporate property insurance", \\"Just want to know about this corporate property insurance", \\"You are a property insurance company",\\ "Yes, this belongs to property insurance "Right?", "Enterprise property insurance", \\"These all belong to property insurance, right?", \\"What about property insurance workers?", \\What do I need if I want to buy this \\insurance for my private enterprise? Procedure","Property Insurance", \\"Well, what does corporate property insurance cover?"}  \\ \hline 
Bad    & \thead{"就是连连续，就是一直一直保", "但是它连不上是怎么回事啊？",\\ "哦就是主要是直接给公司转账对吧？", "接吗", "接也是吗？",\\"我直>接去", "你是直接直接用那个"\\, "就是从哪接过来再接回去","还是需要从哪儿连线这个宽带", \\"你直接给我说这些啊", \\"直接把", "哦直接", "嗯那个礼品是直接就发放了",\\"是你直接给我回复对吗", "直接到那里去",\\ "嗯，最好直飞。", "请帮我连接+", \\"把钱打过去的话，我是直接打到那个证券公司", "直接就是"}& \thead{"It's continuous, it's always guaranteed", \\"But what's wrong with it not being able to connect?",\\ "Oh, it's mainly to transfer money directly to the company, right?",\\ "Yes", "Yes too?" ?",\\"I'll go directly", "Are you using that directly"\\, "Just connect it from where you are and then connect it back",\\ "Or do you need to connect to the broadband from somewhere", \\"You Just tell me this directly", "Just give it directly", "Oh directly",\\ "Well, the gift was given out directly",\\"You replied to me directly, right?","Go there directly" ,\\ "Well, it's best to fly directly.", "Please help me connect +", \\"If I call the money, I will call the securities company directly", "Directly"}\\ \hline
\end{tabular}%
}
\caption{Example of ``Good'' and ``Bad'' intent clusters in coherence evaluation.}
\label{tab:Examples of GOOD/BAD Text Clusters}
\end{table*}
\end{CJK}
\begin{CJK}{UTF8}{gkai}
\begin{table*}[!t]
\centering
\resizebox{\textwidth}{!}{%
\begin{tabular}{|l|l|l|}
\hline
\thead{\textbf{Cluster Name}} &  \thead{\textbf{Original Sentences}} & \thead{\textbf{English Translation}} \\ \hline
\makecell{\textbf{询问-优惠} \\(Inquire-Promotion)} &\thead{'那有什么优惠券什么之类的吗？',\\ '是是怎么形式是优惠吗？',\\'还是不是优惠活动，是那个直接给我打我卡里吗\\，还是是什么优惠券儿啊？',\\ '就比如新用户他有什么优惠券儿之类的吧！',\\ '嗯你你那还有什么优惠的活动吗？\\就是比较合适就是合适的动。'}&\thead{'Are there any coupons or anything like that? ',\\ 'What is the form of the discount? ',\\'Still, it's not a promotion. Is it the one that directly charges my card?\\, or is it some kind of coupon? ',\\ 'For example, what coupons does a new user have? ',\\ 'Well, do you have any other discounts? \\It is more appropriate and appropriate to move. '}\\ \hline
\makecell{解答-金额\\(Answer-Amount)} &\thead{'一共是三十一块二毛',\\ '就每个月一百三十八',\\ '对，一个月也就是四百百四五百块钱嘛，\\给您自己做个积累。',\\ '十二月份的话是用了三十三块九毛二',\\ '对一个月一百三十八'} & \thead{'The total is thirty-one and twenty cents',\\ 'That's one hundred and thirty-eight cents per month',\\ 'Yes, that's four hundred, \\four hundred and five hundred yuan per month. Make an accumulation for yourself. ',\\ 'In December, it cost thirty-three dollars and ninety-two cents',\\ 'That's one hundred and thirty-eight dollars a month'} \\ \hline
\end{tabular}%
}
\caption{Example of cluster naming with "Action-Objective" convention.}
\label{tab:Cluster Naming}
% \vspace{-10px}
\end{table*}
\end{CJK}

\begin{table*}[!t]
\centering \footnotesize
% \resizebox{\textwidth}{!}{
\begin{tabular}{l ccc}
\toprule
\thead{\textbf{Dataset}} & \thead{\textbf{Naming Convention}} & \thead{\textbf{Example (sentence)}} &\thead{\textbf{Example (label)}} \\ \midrule
\textsc{NLU} & \thead{scenario-intent} & \thead{Send an email to Alex and write thank you.} & \thead{email sendemail} \\
\textsc{NLU++} & \thead{list of keywords} & \thead{How long does it usually take to get a new pin?} & \thead{["how\_long","pin",\\"arrival","new"]}  \\ 
% \textsc{ATIS} & \thead{single nonver\textbf{}} & \thead{I would like to find a flight from Charlotte to Las Vegas that makes a stop in st. Louis.} & \thead{flight}  \\ 
\textsc{OOS} & \thead{objective} & \thead{Please tell me why my card was declined yesterday.} & \thead{card\_declined}  \\ \midrule
\textbf{ours} & \thead{action-objective} & \thead{Well, do you have any other discounts?} & \thead{inquire-promotion}  \\ 
\bottomrule
\end{tabular}
% }
\caption{Comparison of cluster naming conventions in existing and proposed intent clustering datasets.}
\label{tab:naming convention}
\end{table*}

\begin{CJK}{UTF8}{gkai}
\begin{table*}[!t]
\centering
\resizebox{\textwidth}{!}{%
\begin{tabular}{|l|l|l|}
\hline
\thead{\textbf{Cluster Name}} &  \thead{\textbf{Clusters with Similar Intention}} & \thead{\textbf{English Translation}} \\ \hline
\makecell{\textbf{询问-意外事故} \\(Inquire-Accident)} &\thead{"假如被车碰了或者是被楼上的砖砸了一下",\\ "给别人儿撞的意外",\\ "啊撞到别人然后就是",\\ "就平时有时候开车嘛可能会遇到这个",\\ "嗯哦这种情况，那要是就是我自己不小心撞到了那个某个地方然后",\\ "就是把其他的东西撞到了呀什么的",\\ "那人生意，不是就是，如果是不小心在马上被车撞了的话",\\ "撞到人了吧",\\ "然后不小在行驶当中被别人损害就是说拿石头砸的呀然后",\\ "什么被车撞了之类的，是吗",\\ "被被被撞了，还是被什么一些什么意外事故了",\\ "我把别人的车撞了是吧"}&\thead{
"If I were hit by a car or hit by a brick from upstairs,"\\
"Accidentally hit by someone else,"\\
"Ah, hit someone else and then,"\\
"Just sometimes when driving, you might encounter this,"\\
"Well, if I'm not careful and hit some place myself,"\\
"Just hit something else or something,"\\
"That's a human affair, not just, if it's accidentally \\hit by a car on the road,"\\
"Hit someone, right?"\\
"Then not small in the process of driving, being \\damaged by someone else, say, hit with a stone, and then,"\\
"What, hit by a car or something, right?"\\
"I hit someone else's car, right?"}\\ \hline
\makecell{询问-意外死亡\\(Inquire-Accident Death)} &\thead{"哦猝死，那猝死算意外吗",\\ "啊，那我想知道那个猝死的话，算是意外死亡吗",\\ "算意外死亡吗",\\ "那如果猝死算是意外死亡吗？",\\ "猝死也算意外事吧",\\ "那那猝死是意外死亡吗如果是猝死的话是",\\ "猝死算是意外死亡吗",\\ "猝死算意外死亡吗",\\ "嗯，那个猝死，猝死属于意外意外险吗，意外死亡吗",\\ "那猝死的话，算意外死亡吗？",\\ "那那那那个就是那个猝死算是意外死亡吗"} & \thead{
"Oh sudden death, is sudden death considered accidental?"\\
"Ah, I want to know if sudden death is considered accidental death?"\\
"Is it considered accidental death?"\\
"If sudden death is considered accidental death?"\\
"Sudden death is also an accident, right?"\\
"Is sudden death considered accidental death if it is sudden death?"\\
"Is sudden death considered accidental death?"\\
"Is sudden death considered accidental death?"\\
"Well, that sudden death, does sudden death fall under \\accidental insurance, accidental death?"\\
"Is sudden death considered accidental death?"\\
"Is that, that, that sudden death considered accidental death?"} \\ \hline
\end{tabular}%
}
\caption{Example of two high-quality intent clusters with similar intentions.}
\label{tab:similar intent}
\end{table*}
\end{CJK}

\begin{table*}[!t]
    \centering
    \begin{tabular}{ccccccc}
        \toprule
        \textbf{Model} & \textbf{Sampling Method} & \textbf{1} & \textbf{2} & \textbf{3} & \textbf{4} & \textbf{5} \\ \midrule
        \multirow{3}{*}{qwen14b} & convex & - & \textbf{96.25}\% & 95.42\% & 95.42\% & 95.21\% \\ \cline{2-7}
        & random (n=10) & \textbf{95.83}\% & 94.38\% & 93.33\% & 92.92\% & 92.92\% \\ \cline{2-7}
        & random (n=20) & \textbf{94.58}\% & 95.00\% & 94.38\% & 94.17\% & 94.17\% \\ \midrule
        \multirow{3}{*}{chatglm3-6b} & convex & - & \textbf{92.08}\% & 90.83\% & 91.04\% & 90.42\% \\ \cline{2-7}
        & random (n=10) & \textbf{92.29}\% & 88.75\% & 85.83\% & 84.58\% & 85.21\% \\ \cline{2-7}
        & random (n=20) & \textbf{90.42}\% & 89.58\% & 89.38\% & 89.58\% & 88.98\% \\ \bottomrule
    \end{tabular}
        \caption{Comparison of sampling methods and hyperparameters for LLM coherence evaluation.}
    \label{tab:sample_method}
\end{table*}

\begin{table*}[!t]
\centering \small
\resizebox{\textwidth}{!}{%
\begin{tabular}{ccccccccccccccc}
\toprule
\textbf{Epoch} & \multicolumn{4}{c}{\textbf{1th}} & & \multicolumn{4}{c}{\textbf{2th}} & & \multicolumn{4}{c}{\textbf{3th}} \\
\cmidrule{2-5} \cmidrule{7-10} \cmidrule{12-15}
& \textbf{n\_cluster} & \textbf{good} & \textbf{bad} & \textbf{rate} & & \textbf{n\_cluster} & \textbf{good} & \textbf{bad} & \textbf{rate} & & \textbf{n\_cluster} & \textbf{good} & \textbf{bad} & \textbf{rate} \\
\midrule
 & 20 & 20548 & 34537 & 0.595 & & 20 & 0 & 2901 & 0.0 & & 20 & 9 & 438 & 0.021 \\
 & 50 & 24100 & 30985 & 0.778 & & 50 & 372 & 2529 & 0.147 & & 50 & 70 & 377 & 0.186 \\
 & 100 & 32292 & 22793 & 1.417 & & 100 & 804 & 2097 & 0.383 & & 100 & 79 & 368 & 0.215 \\
 & ... & ... & ... & ... & & ... & ... & ... & ... & & ... & ... & ... & ... \\
\textbf{Best} & 1600 & 52184 & 2901 & 17.988 & & 800 & 2454 & 447 & 5.490 & & 200 & 86 & 361 & 0.238 \\
\bottomrule
\end{tabular}}
\caption{Example log records from iterative intent clustering across epochs.}
\label{tab:log}
\end{table*}

\end{document}